\RequirePackage{fix-cm}
\documentclass[twocolumn]{svjour3}          

\smartqed  
\usepackage{graphicx}
\usepackage{comment}
\usepackage{amsmath,amssymb} 
\usepackage{color}
\usepackage{subfig}
\usepackage[misc]{ifsym}
\usepackage{amsmath}
\usepackage{cite}
\usepackage{graphicx}
\usepackage{indentfirst}
\usepackage{bbding}
\usepackage{gensymb}
\usepackage{multirow}

\usepackage{amsmath,amssymb,amsfonts}
\usepackage{algorithmic}
\usepackage{graphicx}
\usepackage{textcomp}
\usepackage{amsmath,epsfig}

\usepackage{xcolor}
\usepackage{graphics}
\usepackage{textcomp}
\usepackage{indentfirst}
\usepackage{caption}
\usepackage{epstopdf}
\usepackage{enumitem}
\usepackage{booktabs}

\usepackage{float}

\usepackage{lineno,hyperref}

\begin{document}
\begin{sloppypar}

\title{FT-HID: A Large Scale RGB-D Dataset for First and Third Person Human Interaction Analysis
}


\author{Zihui Guo  \and
        Yonghong Hou \and
        Pichao Wang \and
        Zhimin Gao \and
        Mingliang Xu \and
        Wanqing Li 
}


\institute{Zihui Guo\at
              School of Electrical and Automation Engineering, Tianjin University, Tianjin, China  \\
              \email{gzihui@tju.edu.cn}          
           \and
           Yonghong Hou  \at
            School of Electrical and Automation Engineering, Tianjin University, Tianjin, China  \\
              \email{houroy@tju.edu.cn}          
           \and
           Pichao Wang  \at
            Independent Researcher  \\
              \email{pichaowang@gmail.com}          
           \and
           Zhimin Gao(\Letter) \at
            School of Computer and Artificial Intelligence, Zhengzhou University, Zhengzhou, China  \\
              \email{iegaozhimin@zzu.edu.cn}          
           \and
            Mingliang Xu \at
            School of Computer and Artificial Intelligence, Zhengzhou University, Zhengzhou, China  \\
              \email{iexumingliang@zzu.edu.cn} 
            \and
            Wanqing Li \at
            Advanced Multimedia Research Lab, University of Wollongong, Australia  \\
              \email{wanqing@uow.edu.au} 
}

\date{Received: date / Accepted: date}

\maketitle

\begin{abstract}
 Analysis of human interaction is one important research topic of human motion analysis. It has been studied either using first person vision (FPV) or third person vision (TPV). However, the joint learning of both types of vision has so far attracted little attention. One of the reasons is the lack of suitable datasets that cover both FPV and TPV. In addition, existing benchmark datasets of either FPV or TPV have several limitations, including the limited number of samples, participant subjects, interaction categories, and modalities. In this work, we contribute a large-scale human interaction dataset, namely, FT-HID dataset. FT-HID contains pair-aligned samples of first person and third person visions. The dataset was collected from 109 distinct subjects and has more than 90K samples for three modalities. The dataset has been validated by using several existing action recognition methods. In addition, we introduce a novel multi-view interaction mechanism for skeleton sequences, and a joint learning multi-stream framework for first person and third person visions. Both methods yield promising results on the FT-HID dataset. It is expected that the introduction of this vision-aligned large-scale dataset will promote the development of both FPV and TPV, and their joint learning techniques for human action analysis. The dataset and code are available at \href{https://github.com/ENDLICHERE/FT-HID}{here}.
\end{abstract}


\begin{table*}
	\centering
	\caption{Comparison of FT-HID dataset and other public datasets for RGB-D action recognition}
	\setlength{\tabcolsep}{1.0mm}{
	\begin{tabular}{ccccccccc}
		\toprule
		Dataset&TPV&FPV&Total-Action& Interaction&Subject&Camera&Video-Sample&Year\\
		\midrule
		WCVS dataset \cite{moghimi2014experiments}&\XSolid&\Checkmark&18&0&4&1&918&2014\\
		Robot-Centric Activity dataset \cite{xia2015robot}&\XSolid&\Checkmark&9&9&8&1&189&2015\\
		First-Person Hand Action dataset \cite{garcia2018first}&\XSolid&\Checkmark&45&0&6&1&1,175&2018\\
		EgoGesture dataset \cite{zhang2018egogesture}&\XSolid&\Checkmark&83&0&50&1&24,161&2018\\
		THU-READ dataset \cite{tang2018multi}&\XSolid&\Checkmark&40&0&8&1&1,920&2018\\
		MSRDailyActivity3D dataset \cite{wang2012mining}&\Checkmark&\XSolid&16&0&10&1&320&2012\\
		UTKinect dataset \cite{xia2012view}&\Checkmark&\XSolid&10&0&10&1&200&2012\\
		G3D dataset \cite{bloom2012g3d,bloom2013dynamic}&\Checkmark&\XSolid&20&0&10&1&600&2012\\
		Falling Event dataset \cite{zhang2012rgb}&\Checkmark&\XSolid&5&0&5&1&150&2012\\
		SBU dataset \cite{yun2012two}&\Checkmark&\XSolid&8&8&7&1&300&2012\\
		Osaka dataset \cite{mansur2012inverse}&\Checkmark&\XSolid&10&0&8&1&80&2012\\
		WorkoutSU-10 dataset \cite{negin2013decision}&\Checkmark&\XSolid&10&0&12&1&1,200&2013\\
		Multiview 3D Event dataset \cite{wei2013modeling}&\Checkmark&\XSolid&8&0&8&3&3,815&2013\\
		MAD dataset \cite{huang2014sequential}&\Checkmark&\XSolid&35&0&20&1&1,400&2014\\
		HARL dataset \cite{wolf2014evaluation}&\Checkmark&\XSolid&10&3&21&-&828&2014\\
		Office Activity dataset \cite{wang20143d}&\Checkmark&\XSolid&20&10&10+&3&1,180&2014\\
		ShakeFive dataset \cite{van2014dyadic}&\Checkmark&\XSolid&2&2&37&1&100&2014\\
		G3Di \cite{bloom2014g3di}&\Checkmark&\XSolid&18&18&12&1&--&2014\\
		N-UCLA dataset\cite{wang2014cross}&\Checkmark&\XSolid&10&0&10&3&1,475&2014\\
		UWA3D dataset \cite{rahmani2014hopc}&\Checkmark&\XSolid&30&0&10&-&600+&2014\\
		TJU dataset \cite{liu2015coupled}&\Checkmark&\XSolid&15&0&20&1&1,200&2015\\
		UTD dataset \cite{chen2015utd}&\Checkmark&\XSolid&27&0&8&1&861&2015\\
		MV-TJU dataset \cite{liu2015single}&\Checkmark&\XSolid&22&0&20&2&3,520&2015\\
		NTU-60 dataset \cite{shahroudy2016ntu}&\Checkmark&\XSolid&60&11&40&3&56,880&2016\\
		M$^2$I dataset \cite{liu2016benchmarking}&\Checkmark&\XSolid&22&9&22&2&1,784&2017\\
		UESTC RGB-D dataset \cite{ji2018large}&\Checkmark&\XSolid&40&0&118&8 +360\degree&25,600&2018\\
		NTU-120 dataset \cite{liu2019ntu}&\Checkmark&\XSolid&120&26&106&3&114,480&2019\\
		FT-HID dataset&\Checkmark&\Checkmark&30&30&109&4&38,364&2021\\
		\bottomrule 
	\end{tabular}}
	\label{table1}
\end{table*}

\section{Introduction}
Human action recognition is a fundamental yet challenging task in the field of computer vision. It has a wide range of practical applications including security surveillance, medical care, description generation, video retrieval, and human-robot interaction. 
According to the different emphasis of the analysis, the problem can be generalized into three pairs of learning problems, namely third person vision (TPV) learning and first person vision (FPV) learning, single-view learning and multi-view learning, as well as single-person action learning and multi-person interaction learning. Among them, TPV is the most common data paradigm in current action recognition research \cite{nca1_zong2021multi,nca2_khowaja2020hybrid,nca3_liu2020spatiotemporal,nca4_hou2021local}. When capturing TPV data, the camera can be seen as an observer. The subjects perform the action spontaneously without the intention to interact with the camera. Usually, the TPV camera is fixed at a distance of the subject to get stable and complete action and background environment information. There are many established RGB-D TPV datasets, such as NTU RGB-D dataset \cite{shahroudy2016ntu}, SYSU dataset\cite{hu2015jointly}, UTD dataset\cite{chen2015utd}, and M$^{2}$I dataset\cite{xu2015multi}. However, these datasets still have limitations in some important aspects, such as the number of subjects, background environment, and the number of action categories, which are crucial for building a realistic yet challenging action recognition dataset. For example, SYSU 3D Human-Object interaction dataset\cite{hu2015jointly} only contains 12 distinct actions performed by four subjects, and the large-scale NTU dataset \cite{shahroudy2016ntu} lacks data taken from the top view. The small number of camera views and less characteristic of the subjects may significantly limit the intra-class variation of actions, while the small set of distinguishable classes can be easily recognized and hinder real-world applications. 

Different from TPV learning, the camera is a part of the performer in the collection of FPV data, which is obtained by the camera mounted on a wearable device of the performer. FPV provides the inherent human-centric perspective of human action and has unique characteristics, including the egocentric motion made by the movement of the head of the performer, the detailed information caused by the short distance from the camera to action target, and the narrow field-of-view of the FPV camera. Compared with TPV RGB-D action recognition, limited research has been conducted in the FPV domain. The main reason is the scarcity of the public available RGB-D egocentric dataset. To the best of our knowledge, only five existing FPV datasets \cite{zhang2018egogesture, moghimi2014experiments, garcia2018first, tang2018multi, xia2015robot} contain both RGB and depth modality. These datasets also have the same limitations as TPV datasets in the number of subjects, action category, and background environment. Moreover, most of the mentioned datasets focus on hand action such as gesture and hand-object interaction. However, the actions completed by other body parts as well as the multi-person interaction is also important in daily life.  

TPV and FPV data have their unique strengths and weaknesses respectively. Generally, TPV data is often stationary or moving smoothly, and stays off the subject, which can get a wide range of perspectives and capture more background information, but it lacks the details of the action. FPV data is close to the real scene that the wearer sees, and can record some details that cannot be seen from TPV. However, it gets a limited field of vision and loses some holistic information. Therefore, it is necessary to combine the two types of visions for better human action understanding and analysis. However, most existing datasets only have one kind of vision. For example, the datasets MSR Action 3D \cite{li2010action}, UTDMHAD \cite{chen2015utd} and NTU RGB-D \cite{shahroudy2016ntu} only contain TPV data, while the datasets ADL \cite{pirsiavash2012detecting}, GTEA \cite{fathi2011learning} and Multimodal Egocentric Activity \cite{song2016egocentric} are pure FPV datasets. Although the dataset CMU-MMAC \cite{fathi2011learning} contains both visions, it lacks the depth modality.  

Single-view learning is the conventional approach and it is widely explored in existing research. As all the data are recorded in a static camera view, the core problem of single-view learning is that it does not consider the difference in appearance and motion characteristics from different camera views which may limit the diversity of action expressions and greatly affects the generalization of the recognition algorithm. Besides, some actions are difficult to recognize from certain viewpoints but easy to recognize from others. Thus multi-view learning, which aims to map features obtained from multiple views into a common feature space to handle the variations in visual appearance is in aid of the action recognition field. A few existing datasets  \cite{wei2013modeling, wang20143d, wang2014cross} set multiple cameras from different views but most of them only contain horizontal views and omit the vertical view. For instance, the view in UESTC dataset \cite{ji2018large} cover 360\degree, but they are all at the same height. However, the data from the top view can provide unique information of the action and is especially important for intelligent video surveillance.

Single-person learning includes recognizing the periodic actions performed by only one subject, like walking and waving, and the person-object interaction such as picking a ball and sending a card. Extensive previous research and datasets  \cite{xia2012view, bloom2012g3d, zhang2012rgb, 9609102} focused on single-person action recognition. However, human interactions such as pushing and handshaking are more typical human activities that occur in public places. Human interaction usually involves at least two individual motions from multiple persons, who are concurrently interrelated with each other. Recognition of complex interactions between multiple persons will be necessary for a number of applications, including automatic detection of violent activities in smart surveillance systems. The human interactions are coordinated in the sense that the movement of one person depends on the movement of the other and vice versa, for example, person A kicks person B, and then person B strikes back. In contrast to single-person motions, the study of interactions between multiple persons could help extend the understanding of human motion. Two of the most limitation of existing interaction datasets \cite{yun2012two, wolf2014evaluation, van2014dyadic} are the shortage of interaction categories and the number of samples in each category.  

From the above analysis, it can be seen that pairs of complementary TPV and FPV data can represent the action more comprehensively. Meanwhile, compared with single-view and single-person learning, multi-view and multi-person interaction learning are more challenging and meaningful. However, as illustrated in Table \ref{table1}, the current datasets cannot meet the above three research demands simultaneously and also remain defects in each of the three learning tasks. Therefore, in this paper, a large-scale RGB-D dataset (FT-HID dataset) for first and third person vision analysis with multiple camera views and diverse multi-person interactions is collected to fill the gap. The dataset contains more than 38K RGB samples, 38K depth samples, and about 20K skeleton sequences. 30 classes of daily actions are designed specifically for multi-person interaction with a wearable device and three fixed cameras. FT-HID dataset has a comparable number of data, action classes, and scenes with other RGB-D action recognition datasets. It is more complex as the data is collected from 109 distinct subjects with large variations in gender, age, and physical condition. More importantly, to the best of our knowledge, it is the first large-scale RGB-D dataset that is collected from both TPV and FPV perspectives for action recognition.

Given the FT-HID dataset, we evaluate several published methods based on both handcrafted and deep learning features on TPV action recognition and FPV action recognition separately with suggested cross-subject and cross-view evaluation criteria. Furthermore, to explore the intrinsic relation between TPV and FPV, the performance of fusion across two visions is also investigated. It is believed that the FT-HID dataset can be used as a benchmark and help the community to move steps forward in the exploration of both FPV and TPV action recognition.

Moreover, two novel deep neural networks are proposed for action recognition. One aims to adequately learn the non-linear structure of heterogeneous representations from different modalities and views and to exploit their complementary characteristics. Specifically, four 3D deep convolutional neural networks are utilized to learn the spatio-temporal information of four streams, i.e., TPV RGB videos, TPV depth sequences, FPV RGB videos, and FPV depth sequences respectively. The other one focuses on exploring the joint representation of skeleton sequences from different camera views. It is realized by integrating channel-selection modules into a two-stream structure.

The remainder of this paper is organized as follows: Section 2 briefly reviews some action recognition datasets and approaches. Section 3 introduces the details of the FT-HID dataset, including its collection, structure, and suggested evaluation criteria. Section 4 describes the two baseline methods. Section 5 reports the experimental results and analysis, and Section 6 concludes the paper.

\section{Related Work}
In this section, a brief overview of the current available RGB-D action recognition datasets is proposed. Then, some handcrafted and deep learning based methods are discussed. For both the datasets and methods, TPV and FPV conditions are introduced separately.   
\subsection{Datasets}

\textbf{TPV Dataset.} In the field of human action recognition, most of the established datasets are captured in TPV. Among all the TPV datasets, the MSRDailyActivity3D dataset \cite{wang2012mining}, UTKinect dataset \cite{xia2012view}, UTD dataset \cite{chen2015utd}, NTU-60 \cite{shahroudy2016ntu}, and NTU-120 \cite{liu2019ntu} are commonly used benchmarks. MSRDailyActivity3D dataset \cite{wang2012mining} consists of 320 samples of 16 action classes, which are performed by 10 actors in two background settings. UTKinect dataset \cite{xia2012view} contains 200 samples collected from 10 subjects by Kinect sensor. It collects 10 classes of daily actions in total. In UTD dataset \cite{chen2015utd}, 8 subjects perform 27 actions in a fixed background environment and obtains 861 samples. These three benchmarks contain only single-person actions and are captured with only one camera at a time. In addition, the number of subjects participating in the performance and the overall sample size are also very limited. NTU provides a larger dataset \cite{shahroudy2016ntu} of 60 actions from 40 distinct subjects. Its new version \cite{liu2019ntu} extends the action category into 120 and the total number of samples is 114,480. Although three cameras are used to capture the same action at the same time in NTU datasets \cite{shahroudy2016ntu,liu2019ntu}, only the horizontal views are included, and the vertical view is missing. In addition to the five benchmarks, all the other datasets also contain one or more limitations, including the small number of action categories, interactions, subjects, camera views, and sample size.

\textbf{FPV Dataset.} The number of available FPV datasets is smaller by an order of magnitude than their TPV equivalents. One of the reasons is the difficulty of data collection. Most existing wearable cameras such as GoPro, Goole Glass, and Looxcie only capture RGB videos. Besides, the size of some RGB-D cameras including Kinect, Intel Real Sense, and ZED are relatively big and need real-time computer control to measure depth range which makes them unsuitable for egocentric data collection. There exist only five RGB-D datasets for FPV action recognition. WCVS dataset \cite{moghimi2014experiments} is the earliest work that opened up the research in RGB-D FPV action recognition. Robot-Centric Activity dataset \cite{xia2015robot} is collected from a Kinect device mounted on top of an autonomous non-humanoid robot. First-Person Hand Action dataset \cite{garcia2018first} focuses on daily hand-object interaction and the RGB-D camera is mounted on the shoulder of each subject. EgoGesture dataset \cite{zhang2018egogesture} is designed for egocentric hand gesture recognition which can be seen as a fine-grained action recognition task. Notably, the subjects are taught how to perform each gesture before data collection, which may decrease the inter-class variation and the reality of the dataset. THU-READ dataset \cite{tang2018multi} contains 1,920 video clips with 40 hand actions. During the data collection, the camera is kept in the same direction as the subject’s eyesight. The most obvious limitation of the current FPV datasets is the small number of interaction classes and sample size.

\subsection{Action Recognition Methods}
\textbf{TPV methods.} Since the RGB videos can maintain color and texture information while the depth maps are invariant to illumination changes, many methods \cite{wang2018cooperative,zhang2018fusing,kong2016discriminative,garcia2018modality,shahroudy2017deep} have been proposed to fuse features extracted from both depth and RGB modalities. The handcrafted descriptors, including HOG and HOF, are widely used in early RGB-D action recognition approaches \cite{kong2015bilinear,kong2017max,liu2015rgb,jia2016low,koperski2016modeling}. In addition, some work focus on designing and capturing suitable features to describe the trajectories \cite{gao2017collaborative,liu2018viewpoint,song2014describing,asadi2017action} and motions \cite{gao2015multi,imran2016human,ijjina2017human} of the action. Specifically, Asadi et al. \cite{asadi2017action} propose multi-modal dense trajectories to describe RGB-D videos and utilize a hybrid median filter as well as inpainting technique to enhance depth videos. Based on the dense trajectory features and human pose representations extracted from RGB and depth videos, Gao et al. \cite{gao2017collaborative} propose a collaborative sparse representation learning model for RGB-D action recognition. Imran et al. \cite{imran2016human} train a four-channel deep convolutional neural network to fuse motion history images and depth motion maps extracted from RGB and depth videos respectively. Besides, binary local representation is adopted in \cite{yu2015structure,yu2016structure} to preserve the pairwise structure with shape constraints through an orthogonal projection matrix. Except for RGB videos and depth maps, skeleton sequences are also involved in some work \cite{kong2019collaborative,hu2018deep,seddik2017human,hu2015jointly} to enhance the recognition performance. However, skeleton-based action recognition without RGB video and depth maps can also achieve satisfactory performance due to the advantages of skeleton sequences with detailed position and motion information, insensitivity to the environment, and low storage space.

Existing studies focus on using skeleton data directly to achieve skeleton-based action recognition mainly targets with RNN \cite{zhang2018fusing,fan2018attention,ben2018coding,wang2018beyond}, CNN \cite{cao2018skeleton,nie2019view,vernikos2019image}, and GCN \cite{shi2019two,shi2019skeleton,li2019actional,li2019spatio,wen2019graph,gao2019optimized}. RNN has a strong ability to model sequential data which is consistent with the intrinsic structure of the skeletal sequence and generally takes skeleton joint coordinates vector as the input of networks. Zhang et al. \cite{zhang2018fusing} enhance the RNN through a universal spatial modeling method, then a multi-stream LSTM with a score fusion technique is proposed to combine multiple geometric features. The CNN-based methods convert the skeleton data into pseudo images or joint trajectory maps for simulating the RGB maps. For example, Cao et al. \cite{cao2018skeleton} use thee arrangement strategy, part-based, chain-based, and traversal-based orders, to convert the skeleton sequences into color images. Yang et al. \cite{yang2018action} take advantage of both CNN and RNN to capture semantic structural information and long-term dependencies. In recent years, GCN generalizes CNN to graph-structured data with graph theory. GCN can efficiently distill information from skeleton sequences. Specifically, Shi et al. \cite{shi2019two} propose a two-stream GCN to model both first-order and second-order information adaptively. Li et al. \cite{li2019actional} build AS-GCN by stacking actional structural graph convolution and temporal convolution to model both spatial and temporal features.

\textbf{FPV methods.} Moghimi et al. \cite{moghimi2014experiments} design a domain-specific feature that roughly encodes the distribution and location of skin pixels in the image. The skin segmentation is the first step of the proposed method which includes both RGB and depth information. Then, a variety of descriptors, such as skin histogram and arms-hands bounding box, are built on the top of the obtained skin pixel segmentation and are sent into the classifier for action recognition. Xia et al. \cite{xia2015robot} select four different descriptors that have shown to perform well in classic activity recognition tasks: 3D optical flow, spatio-temporal interesting points in RGB data, depth data, and body posture descriptors. Moreover, they generate independent motion vectors and attention masks to separate ego-motion from independent motion. Tang et al. \cite{tang2018multi} develop a multi-stream deep neural network (MDNN) to adequately learn the nonlinear structure of heterogeneous representations from RGB and depth 
modality as well as exploit the complementary characteristics in TPV videos. It can preserve the distinctive property for each modality and simultaneously explore sharable information in a unified deep architecture. Moreover, in some research, the methods for TPV action recognition, such as two-stream network \cite{feichtenhofer2016convolutional} and VGG16 \cite{simonyan2014very} for RGB videos and HON4D \cite{oreifej2013hon4d} for depth maps are directly used in RGB-D FPV action recognition.
\begin{table*}[hbt]
	\centering
	\caption{The quantitative characteristics of the FT-HID dataset.}
	\setlength{\tabcolsep}{1.0mm}{
	\begin{tabular}{cccccccc}
		\toprule
		Action Class&\#~Sample&Total-Frame(K)&Avg-Frame&Action Class&\#~Sample&Total-Frame(K)&Avg-Frame\\
		\midrule
		Push hands       & 1288 & 290 & 225 & Thumbs up                  & 1276 & 148 & 116 \\
        Beat hands       & 1297 & 248 & 191 & Pour water                 & 1296 & 256 & 205 \\
        Cross clap hands & 1277 & 209 & 164 & Cheers                     & 1298 & 249 & 192 \\
        Clap hands       & 1286 & 187 & 146 & Pass ball by hand          & 1297 & 273 & 210 \\
        Parry            & 1250 & 169 & 135 & Pass ball by feet          & 1290 & 238 & 185 \\
        Pull             & 1275 & 198 & 155 & Pass an object             & 1291 & 169 & 131 \\
        Wave hands       & 1220 & 139 & 114 & Steal from the back pocket & 1272 & 233 & 183 \\
        Kick             & 1287 & 177 & 137 & Feet-guessing              & 1278 & 257 & 201 \\
        Bow              & 1293 & 154 & 119 & Collide with each other    & 1219 & 219 & 180 \\
        Shake hands      & 1280 & 158 & 123 & Lift the table             & 1296 & 279 & 215 \\
        High-five        & 1257 & 137 & 109 & Wrist wrestling            & 1284 & 263 & 205 \\
        Fist to fist     & 1255 & 133 & 106 & Shuffle the cards          & 1264 & 321 & 254 \\
        Finger-guessing  & 1295 & 187 & 145 & Take cards                 & 1304 & 293 & 225 \\
        Celebration      & 1290 & 189 & 147 & Take out cards             & 1302 & 212 & 162 \\
        Push             & 1270 & 179 & 141 & Touch the pulse            & 1277 & 252 & 197 \\
        \midrule
        All Classes      & 38364 & 6441  & 167 &                          &      &     &    \\
		\bottomrule 
	\end{tabular}}
	\label{table_q}
\end{table*}
\section{The FT-HID Dataset}
This section presents the details and the evaluation protocols of the collected FT-HID dataset. Fig. \ref{dataset_samples} shows some sample frames of the dataset.

\subsection{Action classes}
FT-HID dataset contained 30 interaction categories for both first and third person vision in total. A summary of the quantitative characteristics of the dataset is presented in Table \ref{table_q}. The interactions can be divided into two groups, including 23 positive interactions and 7 negative actions. In particular, positive interactions commonly happen in daily life, while negative actions are abnormal actions that usually tend towards violence. The overall categories are as follows.

\textbf{Positive interactions:} Push hands, Cross clap
hands, Clap hands, Wave hands, Bow, Shake hands, High-five, Fist to fist, Finger-guessing, Celebration, Thumbs up, Pour water, Cheers, Pass ball by hand, Pass ball by feet, Pass an object, Feet-guessing, Lift the table, Wrist wrestling, Shuffle the cards, Take cards, Take out cards and Touch the pulse.

\textbf{Negative actions:} Beat hands, Parry, Pull, Kick, Push, Steal from the back pocket and Collide with each other.

\subsection{Subjects}
109 distinct individuals participated in the dataset collection phase. They chose their partner freely and finally formed 89 two-person interaction groups with consistent ID numbers over the entire dataset. For every action category, each group performed four times and the FPV camera wearer changed in action interval.

\subsection{Scenarios}
We provided two different indoor backgrounds with various complex objects, such as desk, armchair, computer, and air conditioner. Three fixed cameras were used to capture TPV videos and the filming range was settled. The participants could walk freely within the range. We encouraged the participants to perform as naturally as possible. Thus the category name of each action was the only hint, and after hearing the category, participants started to perform those actions according to their intuition.

\subsection{Camera Settings and Modalities}

\textbf{First person vision:} Considering the 
available modality, weight, and volume, Orbbec Persee that provided both RGB videos and depth information was utilized in the FPV video capture. The camera was mounted on the helmet and the wearer can move effortlessly as usual. The depth maps and RGB videos were collected and stored about 25 frames per second at the resolution of 320 $\times$ 240. Since the videos for each interaction group were recorded continuously, eight participants were invited to assist video segmentation. They revisited RGB videos and pointed out the exact starting and ending frames of each action. As a result, the original 321 videos were segmented into 8,718 samples.

\textbf{Third person vision:} While collecting FPV videos, data from TPV were recorded synchronously. We used three Kinect v2 to capture two different horizontal views and a vertical view for the same action. The front and side cameras were set up at the same distance,1.8 meters away from the performance area, yet from two different horizontal angles: 0\degree and 45\degree. The top camera was about 2.4 meters from the ground. We collected three major data modalities acquired by the deep camera, namely, the RGB videos, the depth maps, and the 3D skeleton sequences. But the skeleton information was defective in the top camera, thus we only provided RGB videos and depth maps for the top view. The RGB videos were recorded in the provided resolution of 1920 $\times$ 1080 and contained 9,882 samples per view. The resolution of each depth map was 512 $\times$ 424. The skeleton sequence consisted of 3D coordinates of 25 major body joints for each tracked human body in the scene. 
\begin{figure*}[htb]
	\begin{center}
		\includegraphics[width=0.90\linewidth]{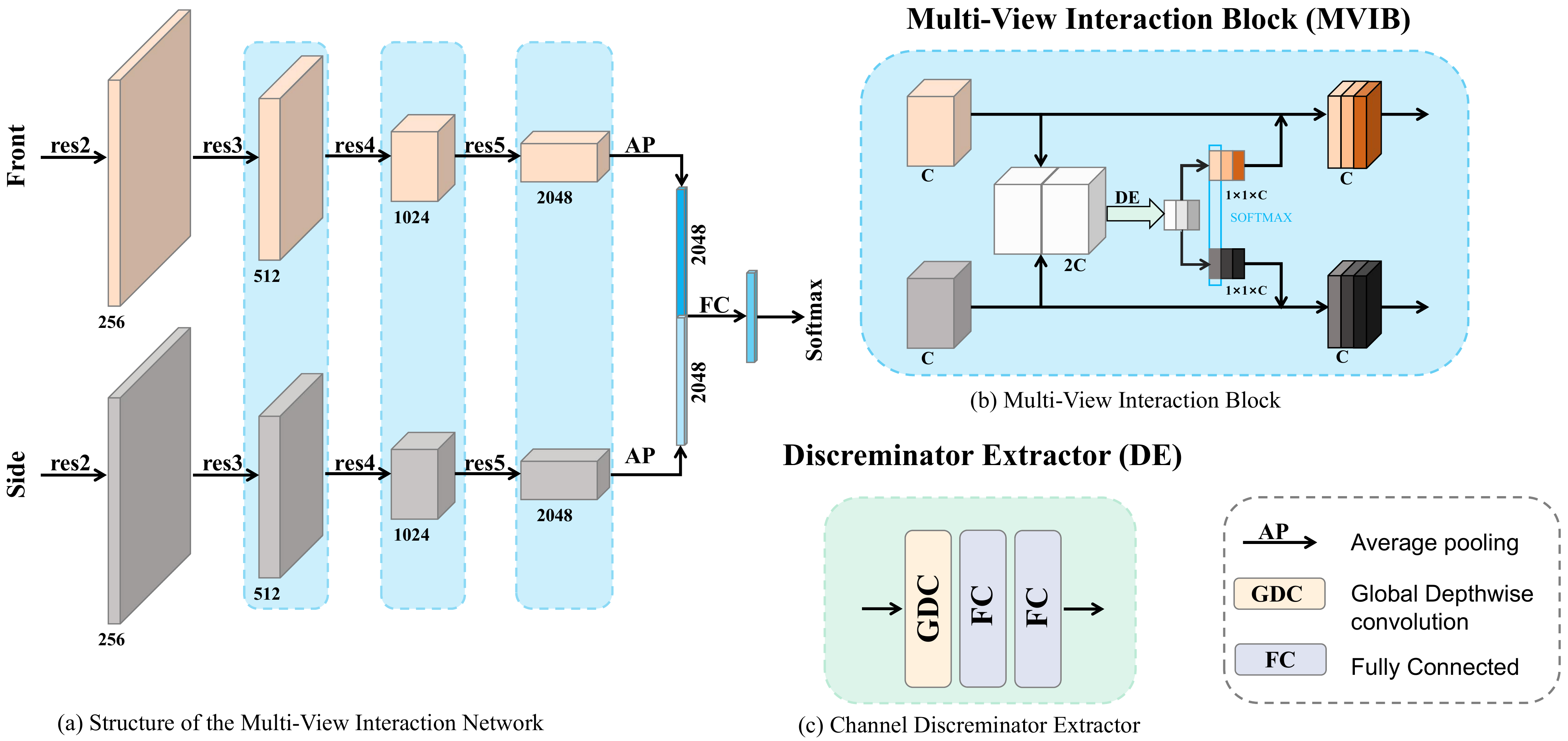}
	\end{center}
	\caption{The overall framework of the proposed multi-view interaction network for skeleton-based action recognition. (a) Structure of the proposed method. Each stream uses ResNet-50 as the backbone, which has four convolution blocks (i.e., res2, res3, res4, res5) to extract features of each view data (i.e., Front, Side). Features from different views interact via MVIB after res3, res4, and res5 respectively, then squeeze the interacted feature from each view via AP, next concatenated these squeezed features, and finally use the concatenated features to predict action category. (b) Illustration of MVIB. (c) Illustration of DE.}
	\label{figure1}
\end{figure*}

\subsection{Evaluation protocols}
In order to fairly evaluate the performances of different methods applied on this dataset, we define several precise evaluation criteria for action recognition. For each criterion, we report the classification accuracy in percentage.

\subsubsection{Cross-Subject Protocol}

In cross-subject (CS) evaluation, 89 interaction groups are split into training and testing groups. For this evaluation, we state two sample split formulation. In the \textbf{First Formulation}, the groups whose IDs are multiples of 4 are utilized as test set and the remaining groups are reserved for training set. In the \textbf{Second Formulation}, the groups whose IDs are before 46 are used as the training data, while the rest of the groups are retained for test data. As the groups whose IDs are before 46 are performed in one background while the remaining groups are in the other background, the second formulation of cross-subject evaluation can also be seen as cross-background evaluation.


\subsubsection{TPV Cross-View Protocol}

For TPV cross-view (CV) evaluation, we choose the side and top camera videos as the training set and the front camera data as the testing set. Specifically, there are 19,764 training samples and 9,882 testing samples.

\subsubsection{TPV and FPV Combined Protocol}

With the help of several participants, we obtain 8,414 paired TPV and FPV (TF) samples. For each view in the four perspectives, we select 5,528 samples and 2,886 samples for training and testing respectively.

\section{Baseline Methods}

In this section, we introduce two new data-driven learning methods to model human actions using the collected 3D action sequences. Specifically, one focuses on the channel correlations in deep models inspired by the current attention mechanism. The other is for paired first and third person vision, which can make full use of the advantages of both perspectives.

\subsection{Multi-View Interaction Network for Skeleton-Based Action Recognition}

To deal with the data from multiple views, some efforts have been made over the past years. Some works \cite{li2020learning_add1,41_zhang2019view} unify the data from different views into a certain coordinate system through rotor and matrix transformation, but the view-specific information might be lost in the process of transformation. Some other works \cite{shao2020learning_add2,wang2018action} utilize parallel networks to extract the features of different views independently, which neglect the correlations among views. However, the view-specific information is beneficial to tackle the occlusions, and the relationships among views can improve the tracking quality. Therefore, considering the distinctions and complementarity of different views, we propose a selective interaction network for multi-view skeleton-based action recognition.


Fig. \ref{figure1} illustrates the overall architecture of the proposed framework, which consists of three steps: imagery mapping, multi-view interaction, and action recognition. Skeleton sequences are obtained from RGB-D cameras. Suppose there are $M$ different cameras observing the same scene from different angles. First, the skeleton sequences obtained from all cameras are mapped into image-like matrices through a body part extension process called imagery mapping. For this mapping, we employ cubic spline interpolation on each of the five body parts as well as a certain joint among all frames, which makes the motion of action smooth and results in a dense representation. Then multi-view interaction is performed using our proposed method which will be described later. And like other deep learning based methods, action recognition is achieved by a fully connected layer together with the softmax activation function.

The interaction strategy of the multi-view action recognition framework can be implemented at different levels: data-level, feature-level, model-level, and decision-level. This work employs the model-level interaction strategy where the interactions are performed in the model learning phase. Moreover, we adopt CNN as the base network due to its outstanding performance in skeleton-based action recognition. 

To sum up, the multi-stream architecture can preserve the characteristics of different views, while the MVIB can capture the complementary information among views. Specifically, the proposed Multi-View Interaction Block (MVIB) is a novel view-interaction mechanism that can integrate into any multi-stream convolutional network for better performance on multi-view skeleton-based action recognition. The details of the mechanism are as follows. Let an action $ A = \{ {A_1},...,{A_m},...{A_M}\} $ be a set of action representations from $M$ different cameras after imagery mapping with $A_m \in {R^{F \times J \times C}}$, where $F$, $J$ and $C$ are the numbers of frames, joints and feature channels respectively. Suppose there are $L$ layers or blocks of the base CNN architecture $Q = \{ {Q_1},...{Q_l},...{Q_L}\} $. Let $A_m^l$ denotes an action feature at layer $l$ from camera $m$. At a multi-view interaction stage, the features from the same layer are concatenated along the channel axis to obtained the integrated feature $I^l$:
\begin{equation}
{I^l} = CONCAT(A_1^l,...,A_M^l)
\end{equation}
Here $CONCAT$ denotes the channel concatenation operation. Then we encode the global information by a global depthwise convolution (GDC) to generate the learnable channel-wise description $d^l$:
\begin{equation}
{d^l} = GDC({I^l}),{d^l} \in {R^{1 \times 1 \times M{C^l}}}
\end{equation}
Further, two fully connected (FC) layers are employed to obtain the correlation representation of all channels:
\begin{equation}
{r^l} = F{C_2}(F{C_1}({r^l})),{r^l} \in {R^{1 \times 1 \times M{C^l}}}
\end{equation}
Note that the output channels of the second FC layer must remain the same as ${d^l}$ for the subsequent selection. Here, ${r^l}$ can be seen as the stack of $M$ vectors which are relative to the features before concatenation:
\begin{equation}
{r^l} = \{ r_1^l,...,r_M^l\} ,{r_m^l} \in {R^{1 \times 1 \times {C^l}}}
\end{equation}
In order to enhance the informative views and weaken the other ones, \emph{softmax} function is used across description vectors of different views. The resulting weight vectors $w^l$ whose value is in the range of 0 to 1, represent the importance of different views in certain channels:
\begin{equation}
{w^l} = \{ w_{\rm{1}}^l,...,w_M^l\}  = \emph{softmax} \{ r_1^l,...,r_M^l\}
\end{equation}
Where ${w^l} \in {R^{1 \times 1 \times M{C^l}}}$. The final feature map $P^l$ is obtained through the weight vectors on various views and the residual connection with the previous feature map: 
\begin{equation}
\begin{aligned}
{P^l} &= {w^l} \cdot {A^l} + {A^l}\\ &= \{ w_1^l \cdot A_1^l + A_1^l,...,w_M^l \cdot A_M^l + A_M^l\}\\  &= \{ P_1^l,...,P_M^l\} ,P_m^l \in {R^{{F^l} \times {J^l} \times {C^l}}}
\end{aligned}
\end{equation}
Then the enhanced feature maps $P_m^l$ of each view are sent into the next convolutional layer for higher-level feature extraction.

The proposed MVIB optimizes the network representation from two aspects. First, for a single view, the weights of different channels indicate their significance for recognizing the action, which achieves channel selection. Second, for multiple views, each view obtains a corresponding importance weight, which contains the correlation with other views, for a certain channel. And the accumulation of these weights along channels demonstrates the contribution of each view to the final classification. Therefore, the proposed mechanism can not only emphasize the vital channels within the interlayer features but also can learn the interaction information between multiple views and guide the network to focus on informative views.

\begin{figure*}[htb]
	\begin{center}
		\includegraphics[width=0.90\linewidth]{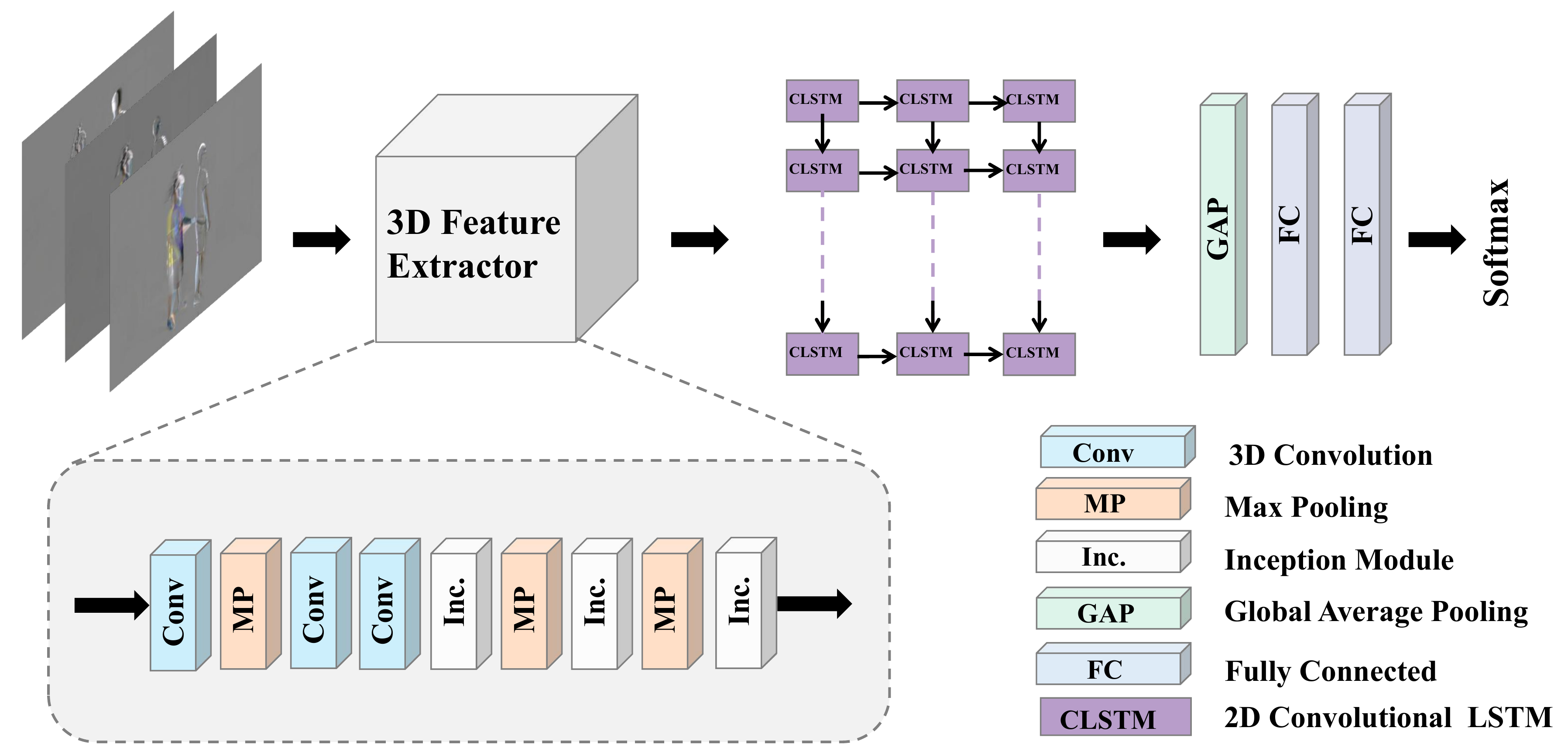}
	\end{center}
	\caption{The overall framework of the proposed SI3D-ConvLSTM Network. The dynamic maps obtained from rank pool operation are firstly sent into a 3D Feature Extractor to learn short-term spatial-temporal features and then three ConvLSTM layers are adopted for long-term spatio-temporal feature extraction. Finally, the high-level spatial-temporal features are used for action recognition through GAP and FC layers.}
	\label{figure3}
\end{figure*}

\subsection{ Dynamic Sequences for FPV and TPV Learning }

In current FPV action recognition methods, video frames are generally directly sent into the neural networks for feature extraction. This ignores the redundant information in the sequence and fails to highlight the distinctive motion patterns. While for TPV action recognition, some works \cite{fernando2016rank,cherian2017generalized,zhu2018rank} compress the initial data through rank pooling to model the video-wide temporal evolution. However, most existing rank pooling based methods generate a single dynamic image for an entire action video, which may lose important sub-action dynamics. By contrast, the proposed method extracts sub-segments using the predefined temporal filters and generates multiple dynamic images corresponding to these sub-segments. Moreover, in order to capture both short-term and long-term spatio-temporal features of the compressed images, we propose SI3D-ConvLSTM Network. The generation of sub-segment dynamics and detailed network architecture are as follows.
\begin{table*}
	\centering
	\caption{The results of different methods, which are designed for depth-based human action analysis, using the two formulations of the CS evaluation criteria on our constructed FT-HID dataset.}
	\setlength{\tabcolsep}{1.5mm}{
	\begin{tabular}{c|cccc|cccc}
		\toprule
		&\multicolumn{4}{c|}{First Formulation}&\multicolumn{4}{c}{Second Formulation}\\
		\midrule
		Method&Front View&Side View&Top View& First View&Front View&Side View&Top View& First View\\
		\midrule
		HOPC \cite{rahmani2014hopc}&0.5132&0.4971&0.5559&0.4798&0.4632&0.4420&0.5007&0.4501\\
		HON4D \cite{oreifej2013hon4d}&0.6135&0.1864&0.5988&0.1957&0.5903&0.0664&0.4614&0.1901\\
		DMMLBP \cite{chen2015action}&0.5979&0.5885&0.6023&0.5550&0.1484&0.4980&0.5644&0.5158\\
		DMMCNN \cite{wang2015action}&0.7953&0.6070&0.6654&0.2904&0.7122&0.2331&0.5513&0.3660\\
		RankPool \cite{fernando2016rank}&0.7474&0.5037&0.6413&0.4936&0.5720&0.3370&0.6310&0.4503\\
		\bottomrule
	\end{tabular}}
	\label{table_cs}
\end{table*}

\textbf{Sub-segment dynamics.} Our main motivation is to exploit the changing of the behavior of the appearance in the frames over time, hence revealing the motion dynamics in short intervals. This is done by applying a handcrafted temporal filter and a rank pooling mechanism to capture temporal statics and features. 

Concretely, assume a set of $N$ action videos, $V = \{ {V_1},...,{V_n},...{V_N}\} $, a video instance $V_n$ with $F$ frames can be represented by a set of ordered continuous frames, $ {V_n} =  < v_n^1,...v_n^f,...,v_n^F >  $, $v_n^f \in {R^d}$. From these sequences of vectors, we adaptively extract the sub-segments using a defined window size $w$ and stride $s$ for ${t_r} = \{ 1,r + 1,2r + 1,...\} $, where ${t_r}$ denotes the starting frame index of the local time interval, i.e. $\{ [t_r^1,t_e^1],...,[t_r^H,t_e^H]\} $. Then the resultant sub-segments $G = \{ {g^h}\} _{h = 1}^H$ are sent into rank pooling module to get dynamic representations.
In each sequence, the number of sub-segments $H$ is fixed to fit the network thus the window size $w$ and stride $s$ varies with the sequence lengths as follows:

\begin{equation}
\left\{ {\begin{array}{*{20}{c}}
	{w = F - (H - 1),F < \varepsilon }\\
	{w = floor(F/H),{F \ge \varepsilon }}
	\end{array}} \right.
\end{equation}
\begin{equation}
\left\{ {\begin{array}{*{20}{c}}
	{s = 1,F < \varepsilon }\\
	{s = w,{F \ge \varepsilon }}
	\end{array}} \right.
\end{equation}
Where $\varepsilon $ is a predefined threshold that divides long and short action sequences. 
The temporal order has shown to be an important feature for video representation as it can effectively model the evolution of the frame appearance. Therefore we propose to encode the sub-segment dynamics by employing a rank pooling algorithm on the segments obtained from the temporal filter. Specifically, we are interested in getting a ranking machine to model a regression space that explicitly imposes a strict criterion that earlier frames must precede the following frames. If ${g^{h+1}}$ succeeds ${g^{h}}$ we have an ordering denoted by ${g^{h+1}} \succ {g^h} $. To encode the dynamics of a sub-segment, the ranking function ${\Psi}({g^h};u) = u^T \cdot g^h$ parametrized by vector $u$. As a result,the aim is to learn the parametric vector $u$ such that it satisfies all constraints $ {\forall {g^a,g^b}},  {g^a} \succ{g^b} \Leftrightarrow {u^T \cdot g^a} > {u^T \cdot g^b}$. Using the principle of RankSVM \cite{joachims2006training} for structural risk minimization and max-margin formulation, the objective function can be expressed as:
\begin{equation}
\begin{array}{l}
\mathop {\arg \min }\limits_u \frac{1}{2}{\left\| u  \right\|^2} + C\sum\limits_{\forall a,b,{g^a} \succ {g^b}}^{} {{\varepsilon _{ab}}} \\
s.t.{u^T}.({g^a} - {g^b}) \ge 1 - {\varepsilon _{ab}}\\
{\varepsilon _{ab}} \ge 0
\end{array}
\end{equation}
To estimate this linear function and learn the parametric vector $u$, we use the Support Vector Regressor (SVR) for its efficiency. The learned vector $u$ defines the pooled feature representation for each sub-segment.

\textbf{Global spatio-temporal extraction and classification.} After the sub-segment dynamics have been encoded, neural networks are adopted to model the global spatio-temporal information of the entire video. ConvLSTM is first proposed in \cite{xingjian2015convolutional}, which can not only establish temporal relations like LSTM but also depict local spatial features like CNN. 



Many experiments proved that ConvLSTM is more effective than LSTM and CNN in obtaining spatio-temporal relations. Therefore, we employ ConvLSTM on the resultant dynamic maps for potent feature extraction. Moreover, to reduce parameters of the network and improve computation efficiency, we design a 3D feature extractor to reduce the dimension of the input sequence and extract short-term spatio-temporal features before three ConvLSTM layers. Specifically, the 3D feature extractor consists of several 3D convolution layers, Inception modules \cite{carreira2017quo} and pooling layers. Finally, the learned high-level representations are fed into a global average pooling layer and two fully connected layers followed by a softmax activation function for action classification. The entire framework named SI3D-ConvLSTM Network is shown in Fig. \ref{figure3}.

\textbf{TPV and FPV Joint learning.} In the FT-HID dataset, TPV consists of three different views while FPV is taken from a single view. Both of them contain RGB videos and depth maps. All kinds of initial data are firstly sent into the sub-segment dynamics generation module for short-term temporal relations modeling. 

Since TPV and FPV are recorded synchronously and follow the same action instance annotation strategy, the dynamic maps for different inputs are one-one correspondence. The dynamics are sent into separate networks for independent global spatio-temporal feature extraction. Score fusion mechanisms are used to combine the predictions of the total eight streams and produce the final recognition category of the current action. The fusion process can take advantage of all kinds of data and enhance the performance of action recognition.

\section{Experiments}
In this section, some conventional handcrafted methods and deep learning methods are evaluated based on the suggested cross-view and cross-subject criteria. Then the performance of the two proposed networks for skeleton-based action recognition and paired-vision action recognition are also evaluated respectively.


\subsection{Results of Existing Action Recognition Methods}
\begin{figure}[htb]
	\begin{center}
		\includegraphics[width=0.90\linewidth]{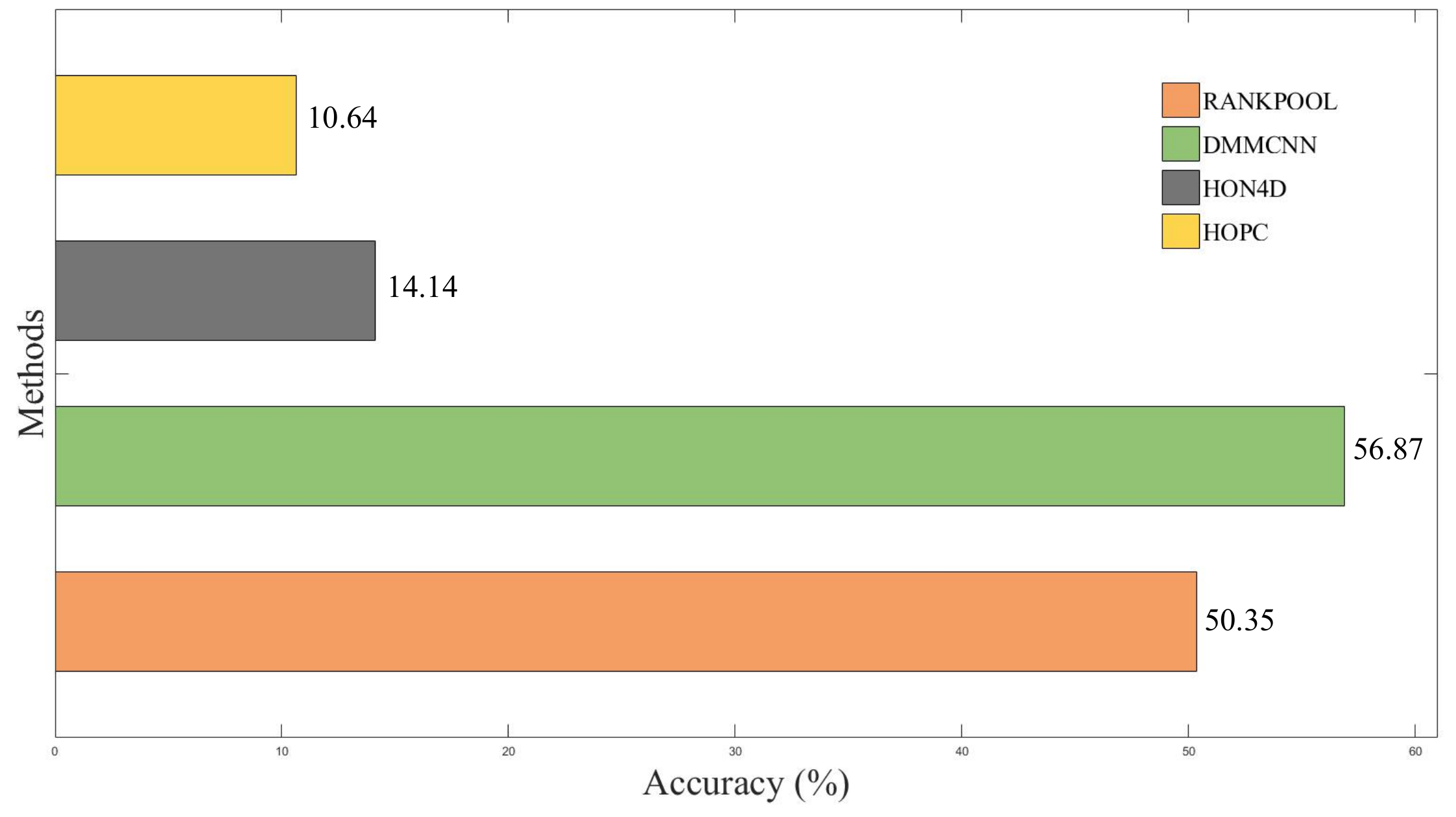}
	\end{center}
	\caption{The performance of different methods on the FT-HID dataset using the CV evaluation criteria.}
	\label{cv_acc}
\end{figure}

We use the publicly available implementation of eight action recognition approaches and apply them to our new dataset. Among them, HOPC \cite{rahmani2014hopc}, 
HON4D \cite{oreifej2013hon4d}, DMMLBP \cite{chen2015action}, DMMCNN \cite{wang2015action}, and RankPool \cite{fernando2016rank} extract features directly from depth maps without using the skeleton information. JDM \cite{li2017joint}, JTM \cite{wang2018action} and IDSF \cite{ding2017investigation} are skeleton-based methods, for which the skeleton data from the front view are used for evaluation. Except the skeleton-based methods \cite{li2017joint,wang2018action,ding2017investigation}, both CS and CV evaluation criteria are used on all the other approaches \cite{rahmani2014hopc,oreifej2013hon4d,chen2015action,wang2015action,fernando2016rank}, and the results are reported in Table \ref{table_cs}, Table \ref{table_cs_skel} and Fig. \ref{cv_acc}. 

\begin{table}
	\centering
	\caption{The results of different methods, which are designed for skeleton-based human action analysis, using the two formulations of the CS evaluation criteria on the FT-HID dataset.}
	\begin{tabular}{ccc}
		\toprule
		Method&First Formulation&Second Formulation\\
		JDM \cite{li2017joint}&0.8603&0.8305\\
		JTM \cite{wang2018action}&0.7632&0.6839\\
		IDSF \cite{ding2017investigation}&0.7788&0.7278\\
		\bottomrule
	\end{tabular}
	\label{table_cs_skel}
\end{table}

As seen, in both CS and CV evaluation, deep learning based methods \cite{wang2015action,fernando2016rank} achieve better results than handcrafted based methods \cite{rahmani2014hopc,oreifej2013hon4d,chen2015action}. In Table \ref{table_cs}, all the results of using the second formulation of CS evaluation are lower than results using the first formulation. It indicates that the second formulation has larger data variation between the training set and the testing set, and is more challenging for existing action recognition approaches. Besides, it can be seen that the result of the front view is the best in most situations because the video captured from the front of the subject can catch more distinctive information than other views. Moreover, the performance of the top view is better than the side view. It reflects the importance of the top view data, which is always be omitted in other action recognition datasets. For some of the methods \cite{rahmani2014hopc,chen2015action}, the recognition results of FPV are comparable to TPV. While for other methods \cite{oreifej2013hon4d,wang2015action}, they are apparently worse than TPV. This reveals that the performance of many existing methods is unsatisfactory for FPV action recognition, and it is urgent to explore methods that can accommodate TPV and FPV simultaneously. The experimental results of the skeleton-based methods under the CS evaluation criteria are summarized in Table \ref{table_cs_skel}. For the first formulation, it can be observed that the three methods outperform the result of the second formulation by 2.98, 7.93, and 5.10 percentage points, respectively. It further indicates the difficulty of the second formulation. Note that the three skeleton-based methods perform better than all the depth-based methods, which verify the efficiency of the skeleton data. JTM \cite{wang2018action} encodes skeleton joint trajectories of each time instance into HSV images, while JDM \cite{li2017joint} employs joint distance as spatial feature with color encoding. IDSF \cite{ding2017investigation} adopts five skeletal features, including joint-joint distance, joint-joint orientation, joint-joint vector, joint-line distance, and line-line angle. Among the three methods, JDM achieves the best recognition accuracy in both formulations and increases the evaluation result by 9.71 and 14.66 percentage points respectively compared with JTM. The results of the CV evaluation criteria are illustrated in Fig. \ref{cv_acc}. HOPC \cite{rahmani2014hopc}, HON4D \cite{oreifej2013hon4d}, and RankPool \cite{fernando2016rank} are hand-crafted methods, while DMMCNN \cite{wang2015action} is a deep learning based method that comprises the weighted hierarchical depth motion maps and a three-channel convolutional neural network. As shown in Fig. \ref{cv_acc}, the deep learning based DMMCNN increases the accuracy of HOPC by 46.23 percentage points and also outperforms other hand-crafted methods by a large margin, which indicates the effectiveness of neural networks.

\begin{figure}[htb]
	\begin{center}
		\includegraphics[width=0.99\linewidth]{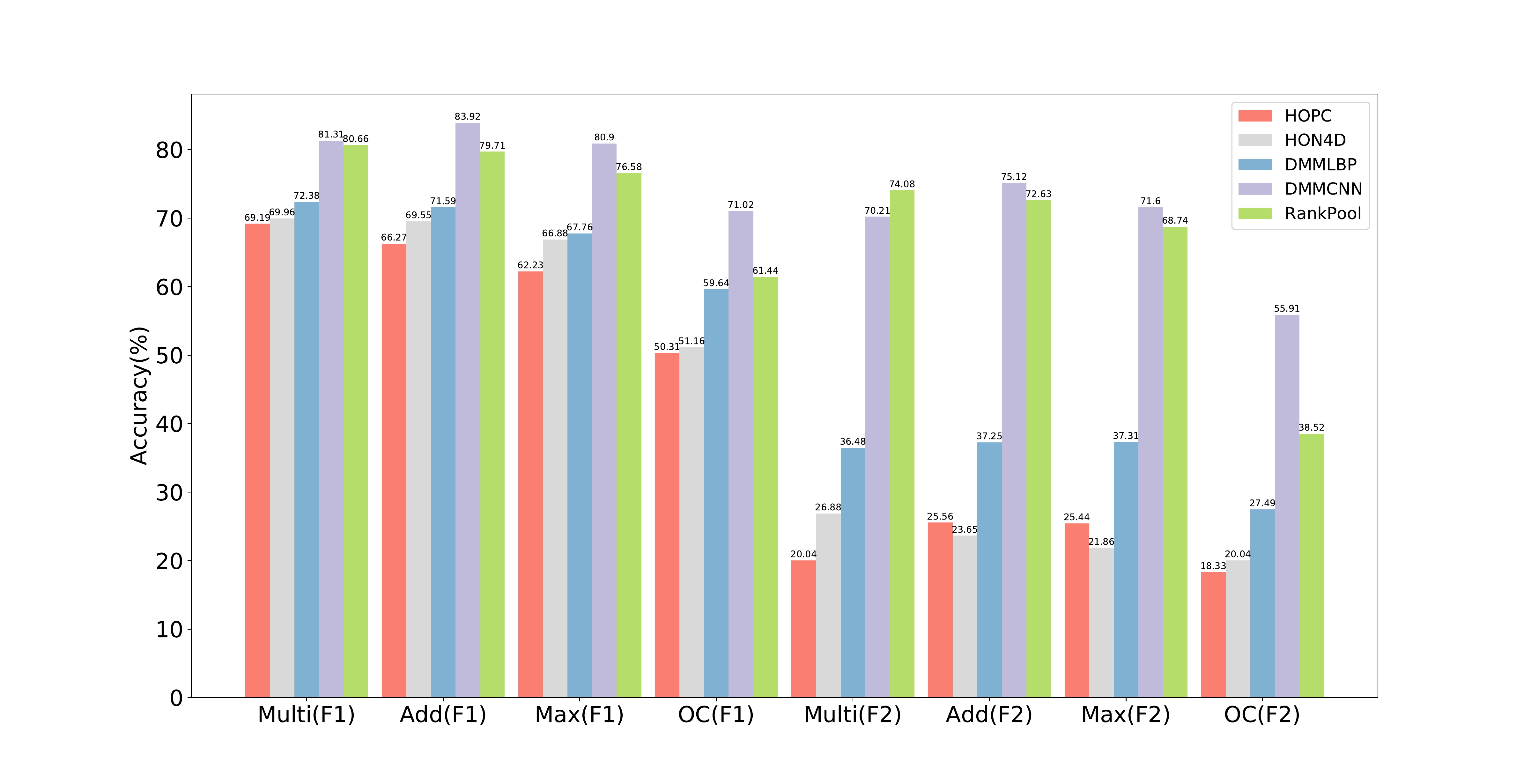}
	\end{center}
	\caption{Comparison of the fusion strategy of TPV with two CS settings. F1 and F2 refer to the first formulation and the second formulation respectively. OC refers to the one channel fusion strategy.}
	\label{table_cs_fusion_TPV}
\end{figure}
\begin{figure}[htb]
	\begin{center}
		\includegraphics[width=0.99\linewidth]{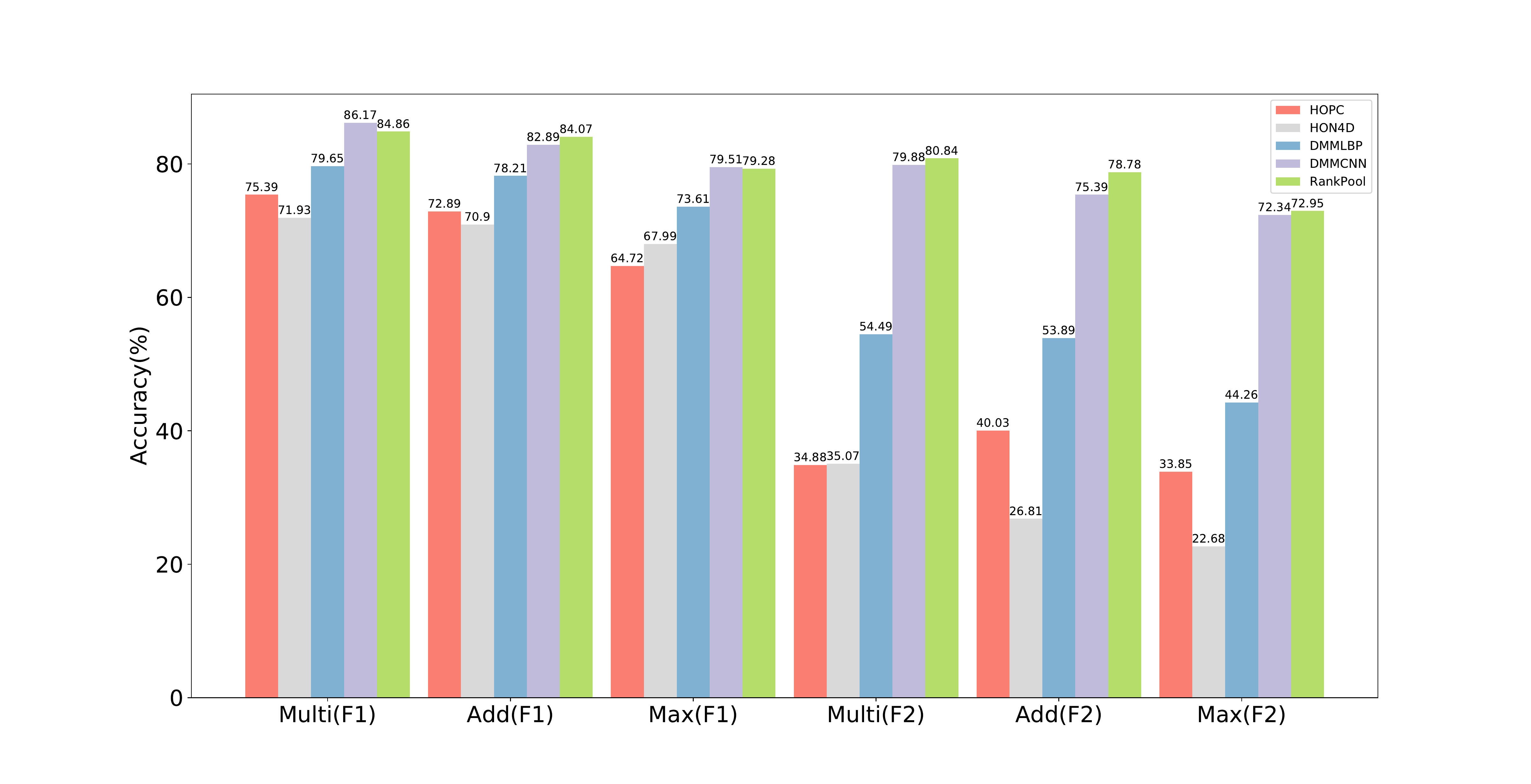}
	\end{center}
	\caption{Comparison of the fusion strategy of TPV and FPV with two CS settings. F1 and F2 refer to the first formulation and the second formulation respectively.}
	\label{table_cs_fusion_TPVFPV}
\end{figure}

Under the CS evaluation criteria, four score fusion strategies, including addition, multiplication, maximum, and a so-called one channel fusion strategy are used to integrate the front, side, and top views in TPV. For the first three score fusion methods, data from different views are trained by three independent networks and the score of each category is fused after the fully connected layer. For the one channel fusion strategy, data from three views are sent them to one network for training. The fusion results of the three views in TPV are reported in Fig. \ref{table_cs_fusion_TPV}. It is clear that the recognition accuracy raises a lot after combining the three camera views but is still restricted by the capability of methods. Meanwhile, the one channel method is the worst among the four fusion strategies. The reason is that the data from the three views are quite different, especially the front and top views. It is difficult for the network to learn effective features when they are treated equally without discrimination. While the separate training can maintain the characteristics of the complementary views and the score fusion improves the performance. Among the four fusion strategies, the multiplicative score fusion in the first formulation consistently outperforms the addition, max score fusion, and one channel methods. This verifies that the feature of the three views is likely to be statistically independent and provide complementary information. However, for some methods, the multiplicative score fusion performs worse than other fusion strategies in the second formulation. It indicates that the design of feature fusion strategy should consider the characteristics and distribution of data.

In order to verify the complementarity between TPV and FPV, the fusion strategy is also applied to these two visions. Considering the characteristics of TPV and FPV are quite different, they are trained separately. Specifically, for the different views contained in TPV, the aforementioned one-channel scheme is used for holistic feature extraction. Then combining TPV and FPV by the score fusion strategy. The results are reported in Fig. \ref{table_cs_fusion_TPVFPV}. From it, we can see that on the evaluated five methods, the integration of two distinct visions significantly improves the performance of the single vision. 
\begin{table}
	\centering
	\caption{The evaluation of MVIB module.}
	\setlength{\tabcolsep}{0.1mm}{
	\begin{tabular}{ccc}
		\toprule
		Method&First Formulation&Second Formulation\\
		\midrule
		concatenation&0.8558&0.6923\\
		score fusion&0.8285&0.5802\\
		MVIB&0.8909&0.7585\\
		\bottomrule
	\end{tabular}}
	\label{table_channel_select}
\end{table}
\subsection{Results of the Multi-View Interaction Network on Skeletons}

We evaluate the performance of the proposed Multi-View Interaction Network on our dataset. Since the FT-HID dataset provides the skeleton data from front and side view, the network in this experiment is a two-stream architecture. The data partition mechanism follows the CS evaluation criteria and both the first and the second formulations are taken into account. To further verify the efficiency of the proposed MVIB module, two contrasting multi-view fusion strategy is designed. In the first contrasting fusion strategy, all the MVIB are removed from the proposed Multi-View Interaction Network, and the final concatenation operation is retained. The second contrasting fusion strategy is to train two independent networks and then adopt multiply score fusion. The results are shown in Table \ref{table_channel_select}. It can be seen that the concatenate fusion performs better than the score fusion, which indicates that using an entire network can better utilize the complementarities among multiple views of skeleton data. Moreover, it is clear that the proposed MVIB module outperforms the other two fusion strategy in both settings, which shows the superiority of the MVIB in multi-view data interaction.

\begin{figure}[t]
	\begin{center}
		\includegraphics[width=0.99\linewidth]{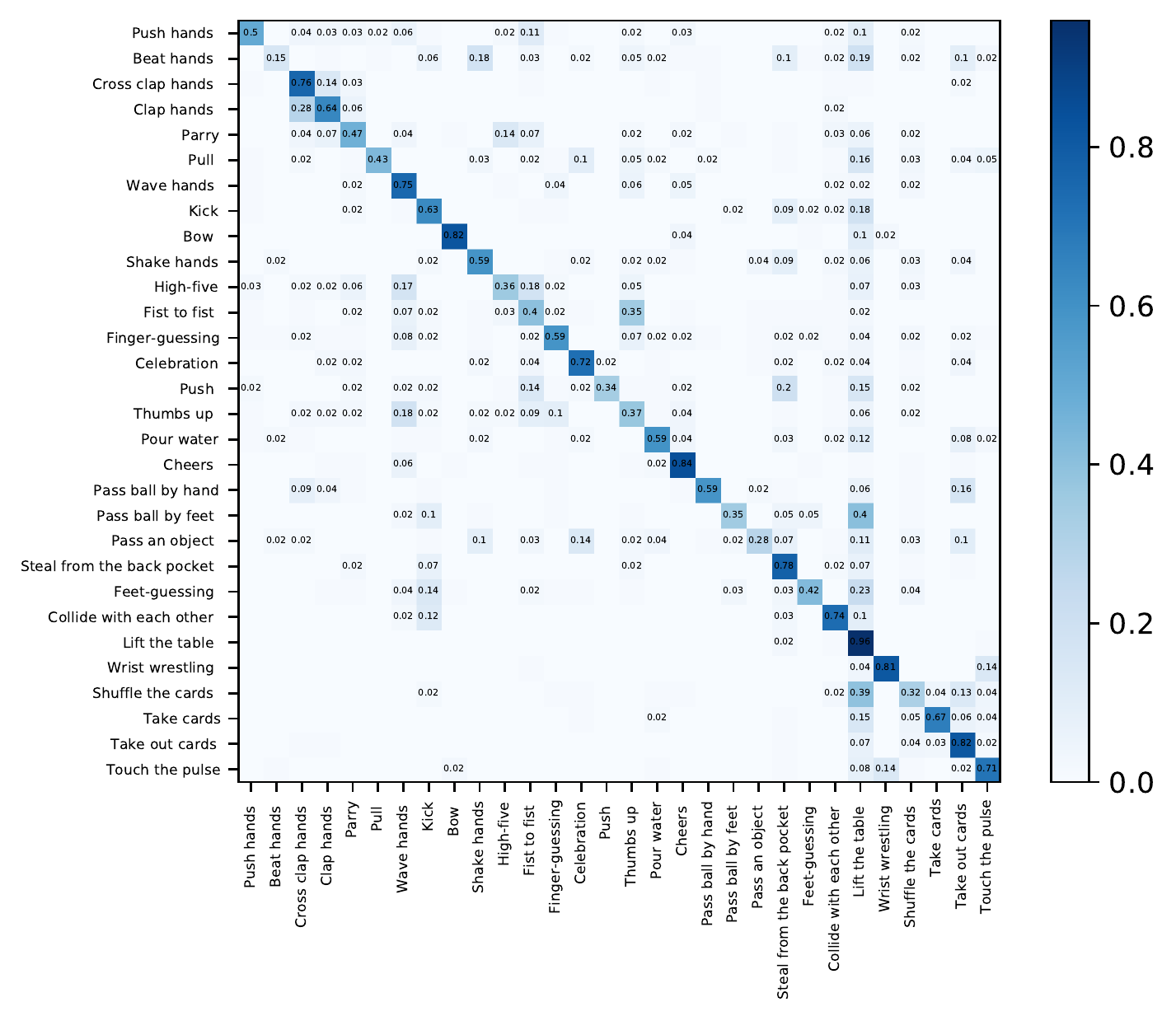}
	\end{center}
	\caption{Confusion matrix of the Multi-View Interaction Network with the second CS formulation.}
	\label{MVIB_Confusion_Matrix_2}
\end{figure}

\begin{table}
	\centering
	\caption{Action recognition results of Multi-View Interaction Network on FT-HID dataset}
	\setlength{\tabcolsep}{2.0mm}{
	\begin{tabular}{cc}
		\toprule
		Top 5 accurate&Top 5 confused pairs\\
		\midrule
		Lift the table&Beat hands$\rightarrow$Lift the table\\
		Cheers&Pass an object$\rightarrow$Celebration\\
		Bow&Shuffle the cards$\rightarrow$Lift the table\\
		Take out cards&Push$\rightarrow$Steal from the back pocket\\
		Wrist wrestling&Pass ball by feet$\rightarrow$Lift the table\\
		\bottomrule
	\end{tabular}}
	\label{top_5_MVBI}
\end{table}

\begin{figure*}[t]
	\begin{center}
		\includegraphics[width=0.90\linewidth]{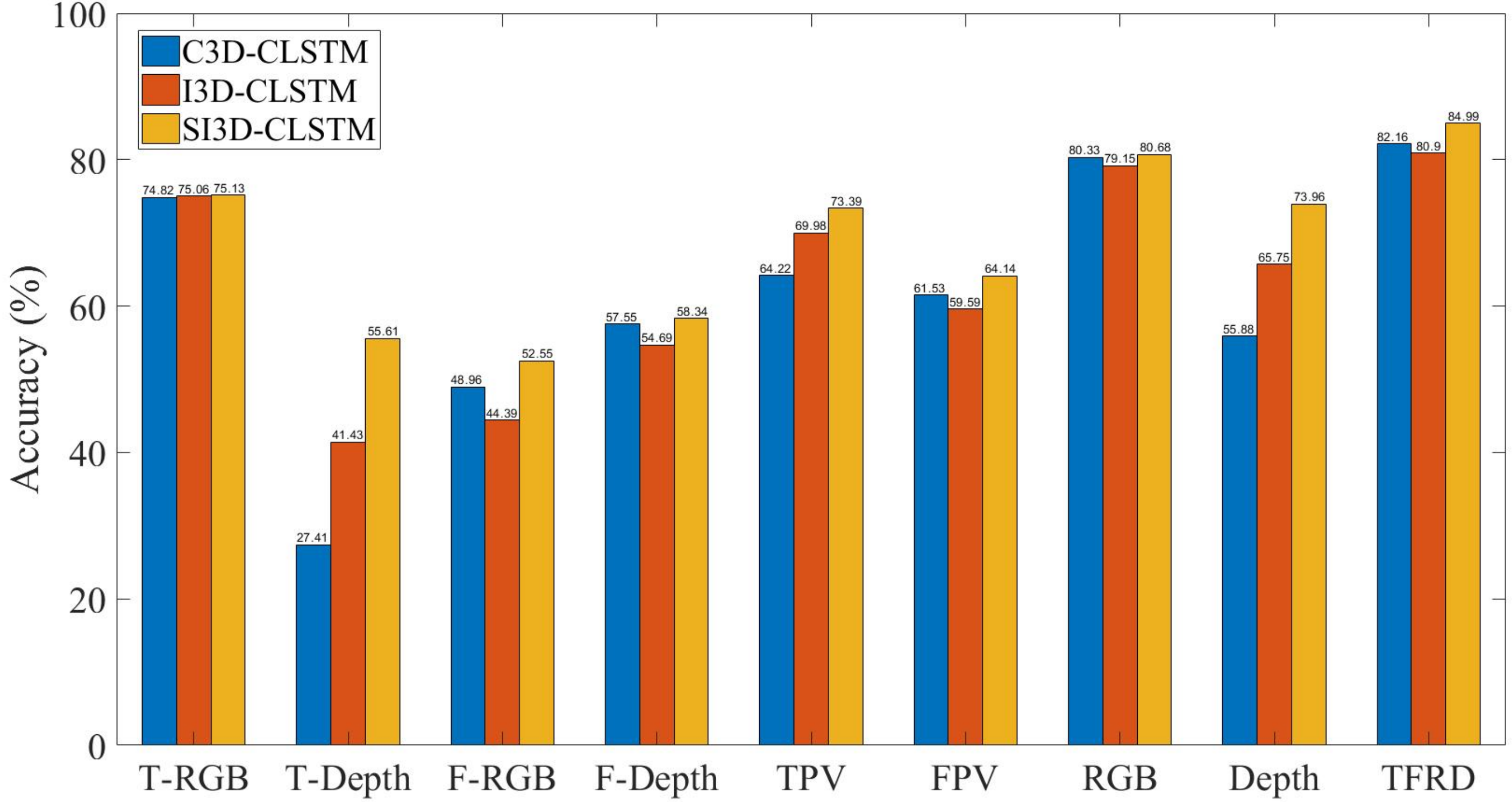}
	\end{center}
	\caption{Comparisons of Accuracy (\%) for the different inputs and the score fusion results on the FT-HID dataset between three 3D-ConvLSTM models. The proposed SI3D-ConvLSTM networks consistently outperform the rest models. Here T and F refer to TPV and FPV respectively, TFRD refers to the fusion of TPV, FPV, RGB, and depth data. }
	\label{clstm_ablation}
\end{figure*}

\begin{figure}[t]
	\begin{center}
		\includegraphics[width=0.99\linewidth]{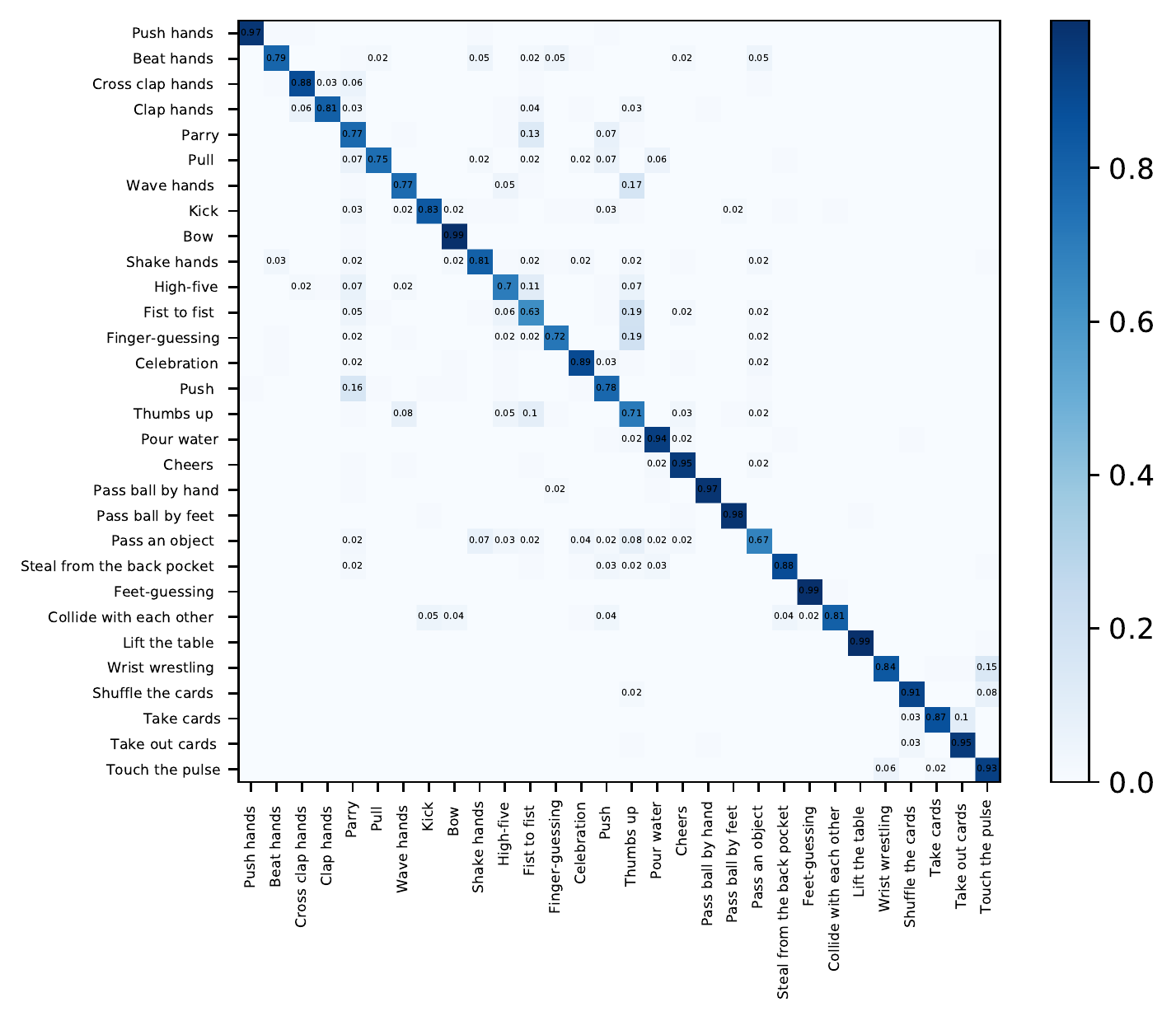}
	\end{center}
	\caption{Confusion matrix of the SI3D-convLSTM Network with the second CS formulation.}
	\label{Confusion_Matrix_ffrd_2}
\end{figure}


The confusion matrix of the proposed network with the second CS formulation is shown in Fig. \ref{MVIB_Confusion_Matrix_2}. To perform in-depth action-wise analysis, action classes that have high recognition accuracy (top 5 accurate classes), and the actions that are easily misclassified to other classes (top 5 confused action pairs) are listed in Table \ref{top_5_MVBI}. Based on the results, it can be found that the actions that have significant motions and discriminative posture patterns could be more accurately recognized. For example, the motion-rich actions “Bow” and “cheers” are in the top 5 accurate actions. Moreover, it can be observed that for skeleton data, the actions that involve interactions with objects may be easily misclassified. For example, “Shuffle the cards” and “Pass ball by feet” are often confused with “Lift the table”, and “Pass an object” is misclassified to “Celebration”. This is possible because the misclassified action pairs have similar human motion patterns, and the perception of the existences of the objects and their appearances is important for accurately recognizing these actions, which can not be learned when using skeleton information only.

\subsection{Results of Dynamic Mapping and SI3D-ConvLSTM Network for TPV and FPV Learning}

We evaluate the performance of the dynamic mapping and the designed SI3D-ConvLSTM network on the FT-HID dataset.  In this experiment, we utilize the TPV and FPV combined protocol. Specifically, the RGB and depth modality of the front view of TPV and FPV are taken into account. The results are reported in Fig. \ref{clstm_ablation}. To verify the effectiveness of the proposed SI3D-ConvLSTM network, two 3D-ConvLSTM networks, I3D-ConvLSTM and Conv3D-CLSTM \cite{zhu2017multimodal}, are also evaluated on the FT-HID dataset. For the fairness of comparison, three networks have the same ConvLSTM layers and fully connected layers, while the spatial pyramid pooling layer in Conv3D-CLSTM is removed. The first four groups in Fig. \ref{clstm_ablation} show the results of the single view and single modality data, including TPV RGB (T-RGB) data, TPV Depth (T-Depth) data, FPV RGB (F-RGB) data, and FPV Depth (F-Depth) data. The intermediate four groups depict the fusion result of each two of the four single inputs, for example, the RGB is the combination of T-RGB data and F-RGB data. The last group illustrates the fusion of all the four single inputs (TFRD). It is clear that in both single input and fusion conditions, the proposed SI3D-CLSTM achieves the highest accuracy and shows its superiority. The confusion matrix of the proposed network in TFRD fusion condition with the second formulation of CS evaluation criteria is presented in Fig. \ref{Confusion_Matrix_ffrd_2}.

Similar to the analysis of the Multi-View Interaction Network, the top 5 accurate actions and the top 5 confused action pairs of five fusion conditions are listed in Table \ref{top_5_CLSTM}. It can be observed that, in contrast to skeleton data, both RGB and Depth data perform well in capturing the object information, many object-related actions could be accurately recognized. For example, “Pass ball by feet” and “Pass ball by hand” are in the top 5 accurate actions of five conditions. Although RGB and depth data modalities both have a good ability in representing the object-related actions, the actions with the object involved may be misclassified when the same object and similar motion patterns are shared by different actions. For example, “Take cards” tend to be misclassified to “Take off cards”. Compared with the TPV data, FPV data suffers more occlusion which makes it easily misclassified to some actions that have similar local motion. For example, when using FPV data as input, “Parry” and “Push” are confused, “Pass an object” tends to be misclassified to “Celebration”. As a result, the TPV data, FPV data, RGB appearance data, and depth data can all affect the recognition performance of different types of action. Meanwhile, the rational fusion of different views and modalities which provide complementary appearance or geometric information may improve the recognition accuracy. 

\begin{table}
	\centering
	\caption{Action recognition results of SI3D-ConvLSTM Network with different visions, modalities and fusions on FT-HID dataset.}
	\setlength{\tabcolsep}{0.5mm}{
	\begin{tabular}{c|cc}
		\toprule
		&Top 5 accurate &Top 5 confused pairs\\
		\midrule
		&Lift the table&Pass an object$\rightarrow$Fist to fist\\
		&Feet-guessing&High-five$\rightarrow$Fist to fist\\
		TPV&Pass ball by feet&Shake hands$\rightarrow$Thumbs up\\
		&Bow&Wrist wrestling$\rightarrow$Touch the pulse\\
		&Touch the pulse&Parry$\rightarrow$Fist to fist\\
		\midrule
		&Shuffle the cards&Fist to fist$\rightarrow$High-five\\
		&Lift the table&Thumbs up$\rightarrow$Wave hands\\
		FPV&Bow&Parry$\rightarrow$Push\\
		&Wrist wrestling&Push$\rightarrow$Kick\\
		&Pass ball by hand&Pass an object $\rightarrow$Celebration\\		
		\midrule
		&Lift the table&High-five$\rightarrow$Fist to fist \\
		&Bow&Fist to fist $\rightarrow$Thumbs up\\
		RGB&Feet-guessing&Parry$\rightarrow$Push\\
		&Pass ball by hand&Thumbs up$\rightarrow$Fist to fist\\
		&Pass ball by feet&Shake hands $\rightarrow$Beat hands\\			
		\midrule
		&Lift the table&Pass an object $\rightarrow$Thumbs up \\
		&Pass ball by feet&Fist to fist $\rightarrow$Thumbs up\\
		Depth&Feet-guessing&Take cards$\rightarrow$Take out cards \\
		&Bow&Parry$\rightarrow$Fist to fist\\
		&Touch the pulse&Shake hands $\rightarrow$Fist to fist\\			
		\midrule
		&Bow&Fist to fist $\rightarrow$Thumbs up \\
		&Feet-guessing&Pass an object $\rightarrow$Thumbs up\\
		TFRD&Lift the table&High-five$\rightarrow$Fist to fist\\
		&Pass ball by feet&Thumbs up $\rightarrow$Fist to fist\\
		&Pass ball by hand&Finger-guessing $\rightarrow$Thumbs up \\			
		\bottomrule
	\end{tabular}}
	\label{top_5_CLSTM}
\end{table}	

\begin{figure*}[p]
	\centering
	\begin{center}
		\includegraphics[width=0.75\linewidth]{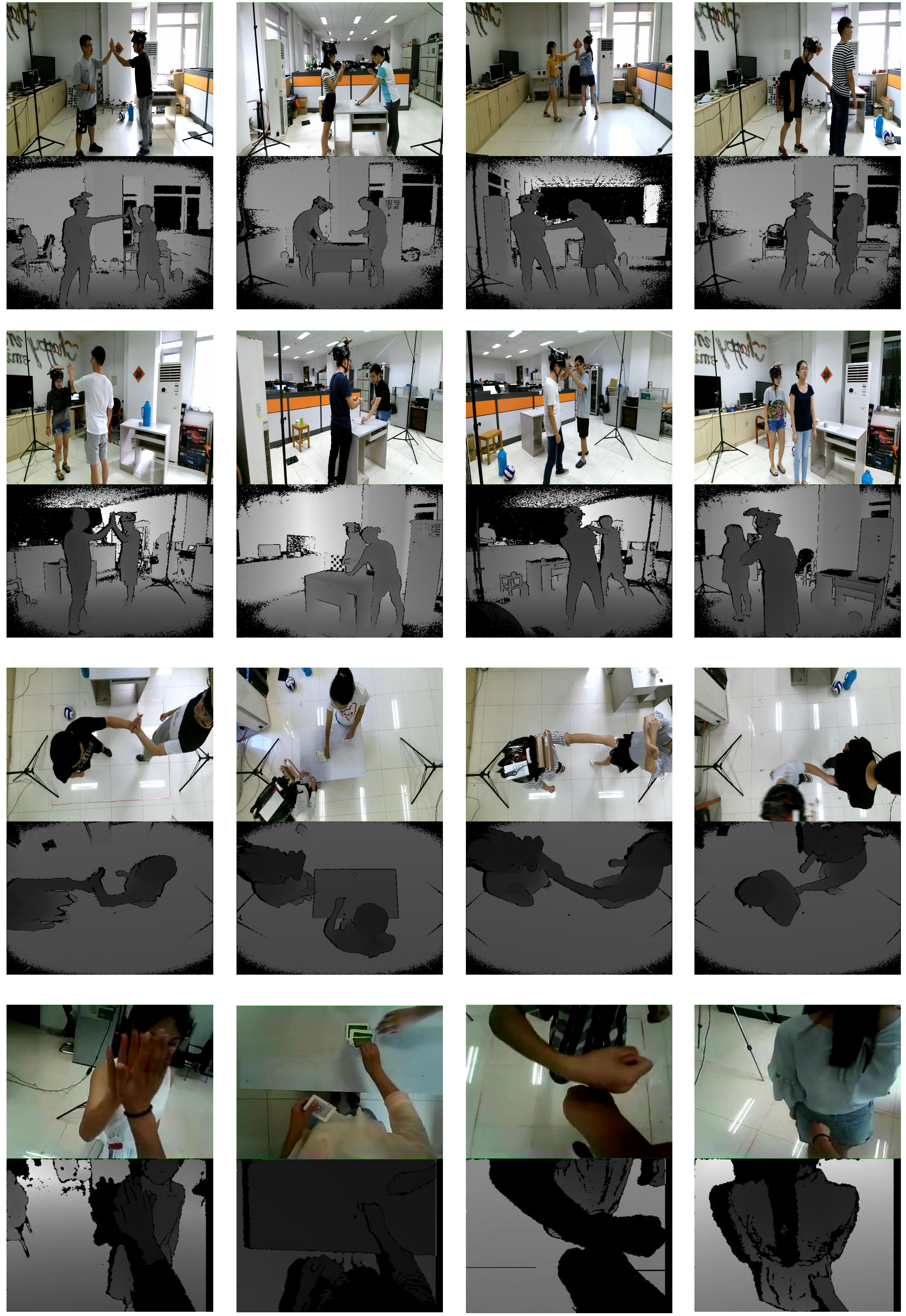}
	\end{center}
	\caption{Samples of RGB and Depth frames of FT-HID Dataset. The four rows present the front, side, top, and first view samples respectively. The first two columns depict two positive interactions (i.e., High-Five and Take Cards), while the rest two rows show two negative interactions (i.e., Parry and Steal From the Back Pocket). These samples illustrate the variety in human subjects, camera views, and environmental conditions.}
	\label{dataset_samples}
\end{figure*}
\section{Conclusion}
A large-scale RGB-D action recognition dataset with pair-aligned TPV and FPV information is introduced in this paper. The dataset has been validated by existing action recognition methods and also by two extended methods. Compared to the existing datasets for the action recognition, the proposed FT-HID dataset has unique features including a large number of samples, distinct interaction categories, diverse camera views, various environments, and a wide range of participants. The FT-HID dataset is expected to facilitate further research in TPV and FPV based human action recognition using individual modalities, multi-modalities, and cross-modalities protocols that covers a broad range of real applications.

%


%
%

\bibliographystyle{spmpsci}      
\bibliography{egbib}   

\end{sloppypar}
\end{document}